\pdfoutput=1
\documentclass[11pt]{article}

\usepackage[]{acl}

\usepackage{times}
\usepackage{latexsym}

\usepackage[T1]{fontenc}
\usepackage[utf8]{inputenc}

\usepackage{microtype}

\usepackage{multirow}
\usepackage{graphics}
\usepackage{array,graphicx}
\usepackage{float}
\usepackage{arydshln}
\usepackage{xcolor}
\usepackage{soul}

\usepackage{bm}
\usepackage{enumerate}
\usepackage{amsmath}
\usepackage{url}
\usepackage{times}
\usepackage{latexsym}
\usepackage{color}

\usepackage{threeparttable}

\newcommand{\mypara}[1]{%

  \paragraph{#1}
}

\title{ATLAS: Improving Lay Summarisation with Attribute-based Control}

\author{Zhihao Zhang$^{1}$, Tomas Goldsack$^{2}$, Carolina Scarton$^{2}$, Chenghua Lin$^{3 \thanks{\quad Corresponding author}}$  \\
        $^{1}$College of Economics and Management, Beijing University of Technology, China,\\
        $^{2}$Department of Computer Science, University of Sheffield, UK \\ 
        $^{3}$Department of Computer Science, The University of Manchester, UK\\
        \small\texttt{zhhzhang@bjut.edu.cn}\quad\texttt{\{tgoldsack1, c.scarton\}@sheffield.ac.uk}\\
        \small\texttt{chenghua.lin@manchester.ac.uk}}

\begin{document}
\maketitle
\begin{abstract}
Automatic scientific lay summarisation aims to produce summaries of scientific articles that are comprehensible to non-expert audiences. However, previous work assumes a one-size-fits-all approach, where the content and style of the produced summary are entirely dependent on the data used to train the model. In practice, audiences with different goals and levels of expertise will have specific needs, impacting what content should appear in a lay summary and how it should be presented. 
Aiming to address this disparity, we propose ATLAS, a novel abstractive summarisation approach that can control various properties that contribute to the overall ``layness" of the generated summary using targeted control attributes.
We evaluate ATLAS on a combination of biomedical lay summarisation datasets, where it outperforms state-of-the-art baselines using both automatic and human evaluations.
Additional analyses provided on the discriminatory power and emergent influence of our selected controllable attributes further attest to the effectiveness of our approach. 
\end{abstract}

\section{Introduction} \label{sec:intro}
Lay summarisation is defined as producing a summary of a scientific article that is comprehensible to non-experts \citep{elifeDigest}. Recent work has shown that, when compared to technical abstracts, lay summaries typically are more readable (lexically and syntactically), more abstractive, and contain more background information, enabling a non-technical reader to better understand their contents \citep{luo-etal-2022-readability, DBLP:conf/acl/CohenKZM20, goldsack-etal-2023-enhancing}. However, the extent to which these attributes are required within a lay summary depends largely on the specific needs of the reader. For example, a scientist from a related field will require less background information to understand an article's contents than an entirely non-technical reader, but they might still require domain-specific jargon to be simplified or explained. Despite its obvious benefits, to our knowledge, no work has yet explored how we can enable such fine-grained control over comprehensibility-related aspects for lay summary generation.

In this paper, we propose \textsc{\textbf{ATLAS}} (\textbf{AT}tribute-controlled \textbf{LA}y \textbf{S}ummarization), a novel scientific summarisation approach that aims to control four attributes targeting distinct properties contributing to the overall ``layness" of the generated summary, thus allowing it to cater to the specific needs of different audiences. 
%both technical and non-technical audiences. 
Although recent attempts at text simplification and story generation have had success influencing the style \citep{martin-etal-2020-controllable,kong-etal-2021-stylized, sheang-saggion-2021-controllable} and content \citep{kong-etal-2021-stylized,tang2024cross} of generated text using fine-grained controllable attributes, no work to our knowledge has explored this for scientific summarisation.  
%Luo et al. 
\citet{luo-etal-2022-readability} recently addressed the task of readability-controlled scientific summarisation, however, this is only done at a binary level, training a model to produce either a technical or non-technical summary based on a single control token. 

Our approach innovates by enabling a greater degree of controllability through the flexible handling of multiple attributes, allowing it to produce more diverse summaries and better address the specific needs of different audiences.
Our results show that ATLAS outperforms state-of-the-art baselines in both automatic and human evaluations across three summary types with varying levels of technicality. Additional analyses confirm that attribute control positively influences performance, and suggest the selected control attributes are able to effectively capture the difference between technical and non-technical summaries.

\section{Methodology} \label{sec:method}
As discussed in \S\ref{sec:intro}, ATLAS aims to control four targeted attributes.
We use BART-base as the base model for ATLAS as it represents the state-of-the-art benchmark in previous lay summarisation works \citep{Guo2020-ba, goldsack-etal-2022-making}.

Formally, each document $x=(x_1, x_2, ..., x_n)$ of length $n$, where $x_i$ is the $i$-th token, is prepended with a control token sequence $l$ such that $x=(l, \bm{x}_{1},\bm{x}_{2}s, ..., \bm{x}_{n})$.
$l$ consists of 
our four selected control tokens, each of which targets distinct characteristics of the output summary that contributes to its overall comprehensibility. 
% (see Figure~\ref{fig:example} for an example sequence $l$ = \textit{L\_1250 R\_12.5 BG\_0.8 CWE\_14}).
We describe each aspect below: 

\mypara{Length (L)} The length of the output summary in characters. A more lay audience may require a longer summary to aid comprehension.

\mypara{Readability (R)} How easy it is to read the text. This is measured using the Flesh-Kincaid Grade Level (FKGL) metric, which estimates the reading grade level (US) required to understand the generated text based on the total number of sentences, words, and syllables present within it.

\mypara{Background information (BG)} The percentage of sentences classified as containing primarily background information. Intuitively, a more lay audience will require greater levels of background information to contextualise an article. 
% Intuitively, increasing this kind of content can aid a reader understanding by contextualising the presented work. 

\mypara{Content word entropy (CWE)} The average entropy of content words. We hypothesise that jargon terms are likely to possess higher entropy values, thus lower average CWE is likely to be a property of more lay text. Since jargon terms are predominately nouns, we extract noun phrases as content words using \textit{CoreNLP} library \citep{manning-etal-2014-stanford}.%Therefore, we extract noun phrases as content words using \textit{CoreNLP} library \citep{manning-etal-2014-stanford}, as jargon terms are predominately nouns. 
We then follow \citet{xiao-etal-2020-modeling} to calculate $I(x_i)$ entropy of a given token $x_i$ as the negative logarithm of its generation probability $P(x_i)$, which is directly extracted from a pre-trained language model. 

\begin{equation}
    I(x_i) = - log P(x_i)
    \label{eq:cwe}
\end{equation}

% Content words for technical text (e.g., jargon terms) are likely to possess higher entropy, thus lower average CWE is likely to be a property of more lay text.

% \vspace{5pt}

During model training, true attribute values (as calculated on reference summaries) are used, allowing the model to learn to associate attribute values with summary properties. 
%At test time, we use average values observed for summaries with different levels of technicality to control each described aspect. 
% Note that it is also possible for users to specify different values for control tokens to generate different types of summaries.
For all attributes, values are discretized into $10$ fixed-width bins depending on their respective range in the train split (from minimum to maximum observed value), resulting in $10$ unique control tokens for each attribute which are added to the vocabulary.
For each attribute at test time, we use the most common bin value observed for reference summaries of the training set as attribute values. 
% For example, to produce eLife-style lay summaries, we provide the model with the bin attribute values that are most common for eLife reference summaries in the eLife train split.

\begin{figure*}[t]
	\centering
	\includegraphics[width=\textwidth]{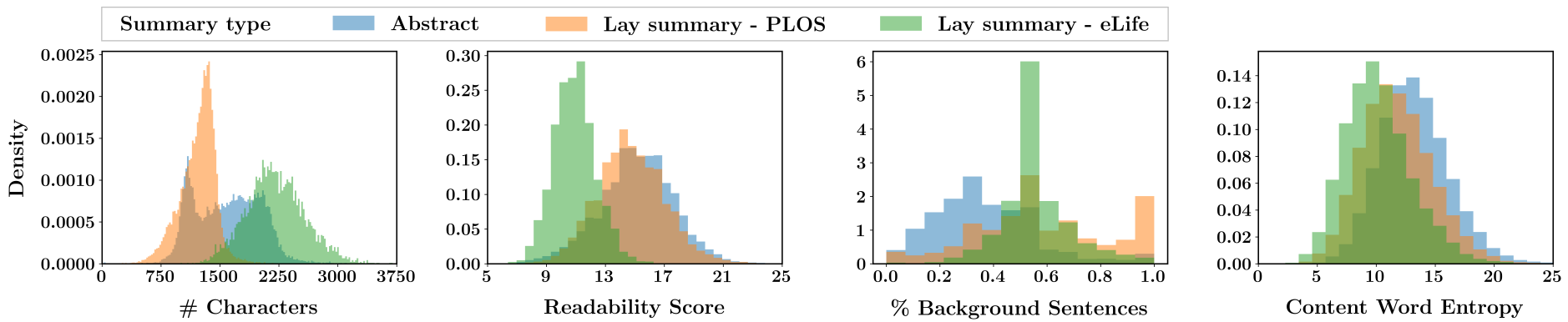}
	\caption{Visualisation of the density distributions of controllable attribute values for each summary type in the combined train split.}
	\label{distribution}
\end{figure*}

\section{Experimental Setup} \label{sec:exp_setup}

\mypara{Data.}
We experiment on the biomedical lay summarisation datasets introduced in \citet{goldsack-etal-2022-making}, eLife (4.8k articles) and PLOS (27.5k articles), for which target lay summaries have been shown to contain different levels of ``layness". Specifically, eLife's lay summaries have been characterized as longer, more readable, and more abstractive than those of PLOS, as well as being empirically observed to be suitable for a more lay audience. We, therefore, combine both of these datasets, allowing us to expose \textsc{ATLAS} to a greater variety of attribute values during training.\footnote{To combine the datasets, we merge the training and validation sets. We evaluate on the test sets separately.} For each article in the combined dataset, we train our \textsc{ATLAS} to produce both the technical abstract and lay summary, using our control attributes to differentiate between them.

\mypara{Evaluation.}
We employ several automatic metrics to evaluate the performance of ATLAS. In line with common summarisation practice, we calculate ROUGE-1,2, and L variants \citep{lin-2004-rouge} and BERTScore \citep{Zhang2019-gi}.
We also measure Dale-Chall Readability Score, a metric that estimates US grade level based on the frequency of common words. 

\mypara{Baselines.}
To enable fair comparison, we rerun many of the baseline approaches used by \citet{goldsack-etal-2022-making} (which have the abstract included in the input) on the combined datasets. Specifically, we rerun the Lead-3, Lead-K, and oracle heuristic baselines; TextRank \citep{mihalcea-tarau-2004-textrank}, LexRank \citep{Erkan_2004}, and HipoRank \citep{Dong2021-yh} unsupervised models; and BART and BART$_{Scaffold}$ supervised models. Here, we use the transformer-based BART base model \citep{DBLP:conf/acl/LewisLGGMLSZ20}, which we fine-tune on our own datasets. 
BART$_{Scaffold}$ is the recreation of a model from \citet{goldsack-etal-2022-making} which is trained using a binary control token (<abs> or <lay>) to produce either an abstract or lay summary for an article. This model is equivalent to that proposed by \citet{luo-etal-2022-readability}, the only previous work on controllable lay summarisation.\footnote{The original code for \citet{luo-etal-2022-readability} is not yet available at the time of writing and their results are reported on a different dataset and thus are not comparable.}

Finally, we include two baselines based on ChatGPT (3.5-turbo), so as to compare against an accessible and widely used method of controlling text generation (i.e., prompt engineering). 
Our first GPT baseline (GPT3.5-zs) uses the following zero-shot prompts: (i) ``\textit{Summarize the following article for an expert audience that is familiar with the technical aspects of the content}'' to generate technical abstracts; (ii) ``\textit{Summarize the following article for a non-expert audience that has some familiarity with the technical aspects of the content}'' to generate PLOS lay summaries, and (iii) ``\textit{Summarize the following article for a non-expert audience that has no familiarity with the technical aspects of the content}'' to generate eLife lay summaries. Our second GPT baseline (GPT3.5-mdc) replicates the method of \citet{turbitt-etal-2023-mdc}, the best-performing team of the recent BioLaySumm shared task \citep{goldsack-etal-2023-biolaysumm}. Based on in-context learning, this method dynamically selects the maximum number of input-output examples that fit in the context window (separated by the simple prompt ``Explanation:'') to generate lay summaries based on only the article abstract. 
% \footnote{Our exact prompts for each summary type will be included in the Appendix of the final paper version.}

\mypara{Implementation Details.}
As mentioned in \S\ref{sec:method}, we employ BART-base as our base model. 
We train our ATLAS for a maximum of $5$ epochs on a GeForce GTX-1080Ti GPU, retaining the checkpoint with the best average ROUGE-1/2/L score on the validation set. 
% We adopt a dynamic learning rate, with warm-up $1000$ iterations and decay afterward. 
We set the batch size to $1$ and keep the $\alpha$ scale factor (\S\ref{sec:method}) at the default value of $0.2$ from \citet{kong-etal-2021-stylized}. 

For calculating control attributes, we use SciBERT \citep{beltagy-etal-2019-scibert} for entropy calculation, and we employ a BERT-based sequential classifier \citep{Cohan2019-ru} trained on the PubMed-RTC dataset \citep{Dernoncourt2017-te} for background sentence classification (as described in \citet{goldsack-etal-2022-making}). We compute the FKGL readability score using the \href{https://github.com/shivam5992/textstat}{\texttt{textstat}} package.

\section{Experimental Results} \label{sec:results}

\begin{table}[t]
	\centering
        \resizebox{0.9\columnwidth}{!}{
	\begin{tabular}{cccc}
	   \hline
	   \textbf{Summary type} & \textbf{Precision} & \textbf{Recall} & \textbf{F1} \\ \hline
	   Abstract  & 0.69 & 0.75 & 0.72  \\ 
	   eLife-Lay & 0.71 & 0.71 & 0.71  \\ 
	   PLOS-Lay & 0.73 & 0.66 & 0.71   \\ \hline
	\end{tabular}
        }
 	\caption{Classifier performance for 3-way classification between summary types on the combined test set.}
        \label{tab:classifier}
\end{table}

\mypara{Discriminatory ability of control attributes.}
To validate the ability of our controllable attributes to distinguish between different summary types, we plot the distribution of attribute values for each type in Figure~\ref{distribution}. The figure suggests that, in combination, the attributes are able to capture characteristic differences between summary types, as instances in which two summary types share a similar distribution for one attribute can typically be separated by other attributes.\footnote{E.g., PLOS lay summaries and abstracts have similar readability distributions but differ in their comprehensibility, length, and entropy distributions. Similarly, PLOS and eLife lay summaries have similar comprehensibility distributions but differ in their readability and length.}

To further evidence this, we use the training set to train a simple logistic regression classifier, using only the attribute values of the reference summaries as features, to discriminate between reference summary types. The test set results in Table~\ref{tab:classifier} show that all summary types are classified with an F1-score above 0.7, attesting to the discriminatory power of our control attributes.

\begin{table*}[t]
    \centering
    \resizebox{1.0\textwidth}{!}{
    \begin{tabular}{llcccccccccccccccccccc} \hline
          \multicolumn{2}{l}{\multirow{2}{*}{\textbf{Model}}} & \multicolumn{6}{c}{\textbf{Abstract}} && \multicolumn{6}{c}{\textbf{Lay summary - PLOS}} && \multicolumn{6}{c}{\textbf{Lay summary - eLife}} \\  \cline{3-8}  \cline{10-15} \cline{17-22} && \textbf{R-1}  & \textbf{R-2} & \textbf{R-L}  & \textbf{BS} & \textbf{DCRS} & \textbf{FKGL} && \textbf{R-1} & \textbf{R-2} & \textbf{R-L} & \textbf{BS} & \textbf{DCRS} & \textbf{FKGL} && \textbf{R-1} & \textbf{R-2} & \textbf{R-L} & \textbf{BS} & \textbf{DCRS} & \textbf{FKGL}   \\ \hline
            % \multicolumn{18}{c}{Heuristics Models}                                            \\  \hline
            \parbox[t]{1mm}{\scriptsize \multirow{3}{*}{\rotatebox[origin=c]{90}{Heuristic}}}
             & Lead-3  & 23.86 & 5.66  & 21.48 & 81.17 & 12.66 & 14.82 && 27.41 & 6.87 & 24.61 & 83.36 & 12.66 & 15.08 && 19.41 & 4.06 & 18.02 & 81.65 & 12.65 & 13.30 \\
             & Lead-K     & 35.69 & 9.07  & 32.70 & 82.86 & 11.69 & 14.49 && 38.28 & 9.45 & 34.8 & 83.72 & 11.88 & 14.95 && 37.27 & 7.53 & 35.18 & 82.05 & 10.58 & 11.89 \\ 
             & Oracle     & 60.08 & 27.48 & 55.95 & 87.35 & 11.12 & 15.15 && 57.82 & 23.92 & 53.37 & 87.13 & 11.20 & 15.28 && 48.92 & 13.42 & 46.30 & 82.94 & 10.51 & 13.18 \\ \hline
            % \multicolumn{18}{c}{Unsupervised Models}                                            \\  \hline
            \parbox[t]{1mm}{\scriptsize \multirow{3}{*}{\rotatebox[origin=c]{90}{Unsup.}}}  & TextRank   & 40.26 & 11.53 & 36.02 & 83.83 & 11.78 & 20.08 && 37.55 & 8.50  & 33.28 & 83.43 & 11.87 & 20.27 && 33.88 & 5.79  & 31.55 & 81.16 & 11.30 & 18.98 \\
            & LexRank    & 38.22 & 13.06 & 35.42 & 83.85 & 9.70 & 14.23 && 31.20 & 9.09  & 28.72 & 82.97 & 9.70 & 14.59 && 32.25 & 5.73 & 30.45 & 80.67 & 9.68 & 13.32 \\
            & HipoRank   & 36.95 & 10.19 & 33.89 & 83.22 & 12.15 & 14.46 && 37.67 & 9.22 & 34.28 & 83.68 & 12.15 & 14.69 && 31.50 & 5.17 & 29.68 & 80.88 & 12.13 & 12.13 \\ \hline
            % \multicolumn{18}{c}{Supervised Models}                                          \\   \hline
            \parbox[t]{1mm}{\scriptsize \multirow{4}{*}{\rotatebox[origin=c]{90}{Supervised}}} & BART & 43.34 & 13.14 & 39.80 & 85.48 & 11.33 & 14.40 && 43.52 & 12.09 & 39.67 & 85.70 & 11.29 & \underline{14.54} && 31.17 & 6.74 & 29.20 & 83.55 & 11.15 & 13.87 \\
            & BART$_{Scaffold}$ & 43.13 & 12.87 & 39.66 & 85.33 & 11.10 & \underline{14.14} && 43.73 & 12.22 & 39.92 & 85.67 & 11.30 & 14.58 && 43.01 & 10.82 & 40.54 & 84.88 & 9.68 & 11.85 \\ 
            & GPT3.5-zs & 28.69 & 6.52 & 15.04 & 82.76 & 11.70 & 14.32 && 42.74 & 12.70 & 22.28 & \underline{86.32} & \underline{10.40} & 13.19 && 33.72 & 8.45 & 16.95 & 84.36 & 10.36 & 13.03\\
            & GPT3.5-mdc & - & - & - & - & - & - && 44.41 & \textbf{14.16} & \textbf{41.12} & \textbf{86.55} & \textbf{10.36} & 13.32 && 37.97 & 9.39 & 35.57 & 84.22 & 10.78 & 13.70 \\  \cdashline{2-22}
            
            & ATLAS & \underline{45.87} & \textbf{14.08} & \underline{42.32} & \underline{85.54} & \textbf{10.96} & 14.21 && \underline{44.44} & 12.33 & 40.60 & 85.70 & 11.22 & 14.58 && \textbf{46.80} & \textbf{12.57} & \textbf{44.14} & \textbf{85.20} & \textbf{8.95} & \textbf{10.87} \\
            
            & ATLAS$_{Oracle}$  & \textbf{46.11} & \underline{14.07} & \textbf{42.51} & \textbf{85.69} & \underline{10.99} & \textbf{14.13} && \textbf{44.97} & \underline{12.49} & \underline{41.02} & 85.82 & 11.21 & \textbf{14.48} && \underline{46.61} & \underline{12.29} & \underline{43.95} & \underline{85.11} & \underline{9.18} & \underline{11.39} \\
            \hline
    \end{tabular}} 
    \caption{Summarization performance on the PLOS and eLife test sets (abstracts combined). R = ROUGE F1 ($\mathbf{\uparrow}$), BS = BERTScore ($\mathbf{\uparrow}$), DCRS = Dale-Chall Readability Score ($\mathbf{\downarrow}$), FKGL =  Flesh-Kincaid Grade Level ($\mathbf{\downarrow}$). For supervised models, we highlight the best score obtained for each metric in \textbf{bold} and \underline{underline} second best. } 
    \label{tab:summ_results_all}
    \vspace{-10pt}
\end{table*}

\mypara{Summarisation performance.}
Table~\ref{tab:summ_results_all} presents the performance of ATLAS and baseline models using automatic metrics on the test sets of PLOS and eLife. We include the results for ATLAS under two conditions: 1) one utilizing the average value for each attribute observed in the training data for each summary type (ATLAS); and 2) one using true attribute values obtained from gold standard summaries (ATLAS$_{Oracle}$), where ATLAS$_{Oracle}$ is intended to provide an upper bound of the obtainable performance using our control attributes.

For all metrics, it is evident from Table~\ref{tab:summ_results_all} that ATLAS exceeds the performance of all baseline approaches for both eLife lay summaries and abstracts, demonstrating a strong ability to control the technicality of generated text whilst producing high-quality summaries. Interestingly, although the GPT3.5-mdc baseline achieves a slightly stronger all-round performance for PLOS lay summaries, it fails to maintain this for the more ``lay" summaries of eLife where ATLAS achieves significantly better performance, indicating that our control attributes can effectively capture these differences.
% \footnote{The performance disparity between datasets for GPT3.5-mdc is likely caused by the relative similarity of PLOS lay summaries to their respective abstracts, as discussed in \citet{goldsack-etal-2022-making}.}

% In most cases, ATLAS$_{Oracle}$ is shown to obtain even greater scores in these metrics, providing further evidence that our selected attributes can successfully capture the stylistic properties of each summary type. 
% Furthermore, ATLAS achieves lower DCRS scores than other supervised models across all summary types, indicating that it consistently produces more readable summaries. 
% Finally, for keyword-based METEOR (KWM), ATLAS models can be seen to consistently obtain the best score, attesting to the model's ability to address the relevant key concepts.
In all cases, ATLAS also achieves scores that are comparable to (and sometimes exceeding) that of ATLAS$_{Oracle}$, suggesting that the use of the most frequently observed bin value for control attributes is effective for producing the appropriate characteristics for each summary type.

\mypara{Ablation study.}
To assess the contribution of each attribute to model performance, we conduct an ablation study, evaluating ATLAS$_{Oracle}$ under different configurations.\footnote{We use ATLAS$_{Oracle}$ as the subject of this experiment rather than ATLAS to get a true reflection of each attribute's influence, rather than an approximation.} Table~\ref{tab:ablation_study} reports the results of this study for abstracts and lay summaries on the combined test sets of PLOS and eLife.

The table shows that the removal of control attributes has a significant detrimental effect on performance. 
% This is somewhat expected, considering that the attribute scores in this experiment contain true values (i.e., obtained from the gold standard summaries), whereas the keyword distributions are predictions.
Additionally, when only a single attribute is included, the length-based control has the highest ROUGE scores, particularly for lay summaries. 
This is to be expected, as lay summaries are known to differ significantly in length between PLOS (avg. 175.6 words) and eLife (avg. 347.6 words).
% , and the predicted summary being of comparable length is known to be beneficial for ROUGE calculation. 
When employing attributes in combination, we can see that the addition of content word entropy control and the subsequent addition of background information control have the greatest benefit to performance for ATLAS with 2 and 3 attributes, respectively. 
Interestingly, no attribute emerges clearly as the least effective as, although readability score control is the only one not included in the 3 attribute model, its inclusion in the single attribute model has clear benefits for lay summary performance. This provides further evidence that, in combination, our control attributes are able to capture the differences between summary types and effectuate them during generation. 

\begin{table}[t]
    \centering
    \resizebox{1.0\columnwidth}{!}{
    \begin{tabular}{lccccccccc} \hline
        \multirow{2}{*}{\textbf{Model}}         & \multicolumn{4}{c}{\textbf{Lay summary}} && \multicolumn{4}{c}{\textbf{Abstract}} \\  \cline{2-5}  \cline{7-10}
                                          & \textbf{R-1}     & \textbf{R-2}    & \textbf{R-L} & \textbf{DCRS} && \textbf{R-1}     & \textbf{R-2}     & \textbf{R-L} & \textbf{DCRS}   \\ \hline
                    
                    BART & 41.68 & 11.29 & 38.12 & 11.27 && 43.34 & 13.14 & 39.80 & 11.33 \\
                    \hdashline
                    +R & 43.34  & 12.03  & 39.75 & 10.91 && 43.49   & 13.23 & 39.95 & 11.12 \\
                    +BG & 42.52  & 11.71  & 39.01 & 11.01 && 43.74   & 13.65   & 40.35 & 10.98 \\
                    +CWE & 41.58  & 11.21  & 38.04 & 11.28 && 44.23   & 13.48   & 40.56 & 11.35 \\  
                    +L & \textbf{44.22}  & \textbf{12.21}  & \textbf{40.55} & \textbf{10.81} && \textbf{44.83}   & \textbf{13.75}   & \textbf{41.31} & \textbf{11.03} \\ \hdashline
                    +L+BG & 44.66  & 12.36  & 40.96 & 10.99 && 45.67   & 13.78   & 42.02 & 11.17 \\
                    +L+R & 44.52  & 12.10  & 40.73 & 10.92 && 45.54   & 13.64   & 41.78 & 11.21 \\
                    +L+CWE & \textbf{44.72}  & \textbf{12.41}  & \textbf{41.04} & \textbf{10.88} && \textbf{45.87}  & \textbf{13.99}   & \textbf{42.32} & \textbf{10.10}  \\ \hdashline
                    +L+R+BG & 44.82  & 12.41  &  41.10 & 10.97 && 45.94   &  14.07  & 42.32 & 11.10 \\
                    +L+R+CWE & 44.83  & 12.39  & 41.05 & 10.90 &&  45.60  &  13.63  & 41.84 & 11.21 \\
                    +L+BG+CWE & \textbf{45.01}  & \textbf{12.56} & \textbf{41.38} & \textbf{10.88} &&  \textbf{46.04}  & \textbf{14.16}   & \textbf{42.44} & \textbf{11.06} \\
                    \hdashline
                    ATLAS$_{Oracle}$    & 45.22   &  12.47 & 41.45 & 10.91 &&  46.11 &  14.07  &  42.51 & 10.99 \\ 
                    \hline
            \end{tabular} 
    }
    \caption{Ablation study on the ROUGE-based performance of ATLAS under different configurations using true attribute values. ``+" denotes aspect addition. L = Length, R = Readability, CWE = Content Word Entropy, BG = Background information.} 
    \label{tab:ablation_study}
\end{table} 

\mypara{Human evaluation.}
To provide a comprehensive assessment of the summaries generated, we conducted a human evaluation involving our proposed model ATLAS and the strongest baseline model (BART) using two experts.\footnote{Both judges have experience in scientific research and hold at least a bachelor's degree.} Specifically, adopting a similar setting to the original that of \citet{goldsack-etal-2022-making}, we take a random sample of 10 articles from the test split of each dataset. Alongside each model-generated lay summary, judges are presented with both the abstract and reference lay summary of the given article. We choose not to provide judges with the full article text in an effort to minimise the complexity of the evaluation and the cognitive burden placed upon them. Using 1-5 Likert scale, the judges are asked to rate the model output based on three criteria: (1) \textit{Comprehensiveness}: to what extent does the model output contain the information that might be necessary for a non-expert to understand the high-level topic of the article and the significance of the research; (2) \textit{Layness}: to what extent is the content of the model output comprehensible (or readable) to a non-expert, in terms of both structure and language; (3) \textit{Factuality}: to what extent is the model generated lay summary factually consistent with the two other provided summaries (i.e. abstract and reference lay summary).\footnote{For example, for the ``Layness“ criteria, a score of 5 is equal to ``highly lay" and a score of 1, ``highly technical".}

\begin{table}[t]
    \centering
    \resizebox{0.9\columnwidth}{!}{
    \begin{tabular}{lccccc} \hline
        \multirow{2}{*}{\textbf{Criteria}} & \multicolumn{2}{c}{\textbf{eLife}} && \multicolumn{2}{c}{\textbf{PLOS}} \\  \cline{2-3} \cline{5-6}
        & \textbf{BART} & \textbf{ATLAS}  && \textbf{BART} & \textbf{ATLAS} \\ 
                \hline
                Comprehensiveness  & $2.30$ & $2.65$ && $2.00$ & $2.55$ \\
                Layness   & $2.60$  & $3.05$  && $2.10$  & $2.45$ \\
                Factuality     & $2.20$ & $2.85$  && $2.05$  & $2.40$ \\
                \hline
        \end{tabular}
        }
    \caption{Human evaluation on eLife and PLOS. Mean evaluator ratings (1-5) obtained by BART and ATLAS outputs for each metric.}
    \label{tab:human_eval}
\end{table}

Table~\ref{tab:human_eval} presents the average ratings from our manual evaluation. We calculate the Cohan Kappa scores to measure inter-rater reliability, where we obtain values of 0.50 and 0.57 for eLife and PLOS, attesting to the reliability of our evaluation. The overall results suggest that our proposed method performs better than the BART baseline in terms of all three criteria on both datasets, attesting to their quality. In terms of layness, the higher layness scores observed in the eLife dataset compared to the PLOS dataset align with the previous analysis for the two datasets from \cite{goldsack-etal-2022-making}. Moreover, compared to baseline, it is worth noting that our model outputs are judged to produce much more factually correct outputs on both datasets, suggesting our method generates fewer hallucinations.

\mypara{Controllability analysis.}
To assess the extent to which our control attributes enable controllability over the overall layness of the text, we conduct a further analysis using two additional versions of ATLAS with highly lay or technical values. Specifically, we create ATLAS$_{lay}$ and ATLAS$_{technical}$ by selecting the lowest and highest attribute bins, respectively, for which there are at least 100 observations in the training data (for all attributes other than length which is kept constant). 

\begin{table}[]
    \centering
    \resizebox{0.9\columnwidth}{!}{
    \begin{tabular}{llccc}
        \hline
         \multicolumn{2}{l}{\textbf{Model}} & \textbf{FKGL} & \textbf{CLI} & \textbf{DCRS} \\
         \hline
         \parbox[t]{1mm}{\scriptsize \multirow{2}{*}{\rotatebox[origin=c]{90}{PLOS}}}&ATLAS$_{technical}$ & 15.11 & 14.21 & 11.64 \\
         &ATLAS$_{lay}$ & 13.22 & 13.97 & 11.22 \\ \hline
         \parbox[t]{1mm}{\scriptsize \multirow{2}{*}{\rotatebox[origin=c]{90}{eLife}}}&ATLAS$_{technical}$ & 14.77 & 14.02 & 11.32  \\
         &ATLAS$_{lay}$ &10.89 & 11.45 & 9.17 \\ \hline
    \end{tabular}
    }
    \caption{Readability metrics for two versions of ATLAS with highly lay and technical attribute values.}
    
    \label{tab:readability_case_study}
\end{table}

We examine how these extreme attributes manifest themselves in generated summaries by calculating the average readability values obtained by the generated summaries for both datasets. We present the results of the analysis in Table \ref{tab:readability_case_study}, which show a significant divergence in the readability values obtained by each model on both datasets. Interestingly, this divergence is substantially wider for summaries generated on eLife, the dataset which is identified by \citet{goldsack-etal-2022-making} as containing lay summaries that are more ``lay'' than those of PLOS, suggesting that exposure to more extreme values whilst training on this dataset may enable even greater controllability at inference time.\footnote{Examples of summaries generated by these models are included in the Appendices.}

\section{Conclusion}
In this paper, we introduce ATLAS, a model for controllable lay summarisation that employs controllable attribute tokens to influence various properties of the generated summary, enabling it to cater to users of different levels of expertise. Using combined datasets for biomedical lay summarisation we perform multiple experiments whereby we confirm the ability of our selected control attributes to discriminate between summary types, demonstrate their effectiveness for controllable lay summarisation, and further investigate their ability to effectuate desired differences during generation.
\section*{Limitations}
\label{sec:limitations}
Although our results demonstrate that our selected control attributes are able to effectively capture the characteristics between summary types, it is highly likely that there are additional attributes that we have not explored that could benefit performance for controllable lay summarisation. We plan to explore this in future work, in addition to experimenting with more complex methods for enabling controllability. 

% Entries for the entire Anthology, followed by custom entries
\bibliography{anthology,custom}
\bibliographystyle{acl_natbib}

\appendix

\section{Appendix} \label{sec:appendix1}

\paragraph{ChatGPT Baseline Prompts}
The prompts provided to ChatGPT for each summary type are given in Table \ref{tab:prompts}. To ensure a fair comparison, we control the length of the GPT baselines using the generation arguments, (e.g., max\_new\_tokens). Note that we differentiate between the lay summary types (namely, PLOS and eLife) based on distinctions made by \citet{goldsack-etal-2022-making}, who recognise PLOS' summaries as the less ``lay" of the two, making them better suited to an audience with some technical knowledge.

\begin{table*}[]
    \centering
    \begin{tabular}{ll}
        \hline
         \textbf{Summary Type} & \textbf{Prompt}  \\ \hline
         \multirow{2}{*}{\textbf{Abstract}} & Summarize the following article for an expert audience that is familiar\\
         &with the technical aspects of the content \\ \hline
         \multirow{2}{*}{\textbf{PLOS lay summary}} & Summarize the following article for a non-expert audience that has some\\ 
         &familiarity with the technical aspects of the content \\ \hline
         \multirow{2}{*}{\textbf{eLife lay summary}} & Summarize the following article for a non-expert audience that has no\\&familiarity with the technical aspects of the content \\ \hline
    \end{tabular}
    \caption{Prompts used for the GPT3.5-zs baseline for each summary type.}
    \label{tab:prompts}
\end{table*}

\begin{figure*}[t]
    \noindent\fbox{%
        \parbox{\textwidth}{%
            \begingroup
                \begin{center}
                \textbf{eLife}\\    
                \end{center}
                
                \fontsize{10pt}{12pt}\selectfont

                \underline{ATLAS$_{technical}$} \\ 
                The effects of muscle fatigue on motor learning under fatigue are poorly understood. Here, we investigated the effect of fatigue on learning under a sequential pinch force task.  Irrespective of whether the observed fatigue effects are domain-specific or present in another task that is cognitive demanding but requires minimal force control, we found that participants had impaired skill learning in both the fatigued and unfatigued effector.  We replicated the findings of experiment 1 and found that disruption of rTMS to the motor cortex (Cantarero et al ., 2013a) alleviated the adverse effects of fatigue.  Cortical excitability was similar to that observed in the untrained effector, but not in the unfatigued. Altogether, our findings suggest that motor fatigue has a domain-dependent lasting effect on skill learning. Future studies should focus on understanding the role of motor cortex excitability in the acquisition of motor skills under fatigue, as well as the potential role for maladaptive memory formation under fatigued conditions. Cortical and motor cortices should be included in training and rehabilitation regimens geared to improve motor skill acquisition.\\
                
                \underline{ATLAS$_{lay}$} \\
                Muscle fatigue is a neuromuscular phenomenon that can impair performance over time. People who experience fatigue tend to be less able to learn a new motor skill than people who experience no fatigue. However, it is not clear how fatigue affects the ability of people to learn new motor skills . One way to study the effects of fatigue is to study how people learn a motor skill under fatigue conditions. One of the main challenges in studying motor learning under fatigue is the so-termed ``performance-learning" distinction  In this study, participants were asked to practice a motor task over two days and then had to catch up to the skill performance level of the non-fatigued group. Unexpectedly , participants who were only fatigued at the end of the training were less likely to learn the motor skill. This suggests that fatigue has a domain-specific lasting effect on the learning of a skill. ernas et al. now show that people who are unable to recover the motor task under fatigue are more likely to be unable to learn their motor skill when they are not fatigued. The experiments show that when people are trained to perform the task, their ability to recover from fatigue is severely impaired. This effect is due to a change in the strength of the motor cortex, a region of the brain that is involved in learning and memory. 
            \endgroup \vspace{4pt}
        }
    } 
    \caption{An case study from the eLife test set comparing summaries generated under highly lay and technical attribute values (with the length attribute being kept constant).}
    \label{fig:case_study_elife}
\end{figure*}

\begin{figure*}[t]
    \noindent\fbox{%
        \parbox{\textwidth}{%
            \begingroup
                \begin{center}
                \textbf{PLOS}\\    
                \end{center}
                
                \fontsize{10pt}{12pt}\selectfont

                \underline{ATLAS$_{technical}$} \\ 
                In this paper, we explore the conditions under which associations between antigenic, metabolic and virulence properties of strains within pneumococcal populations and predict how these may shift under vaccination. In this work , we use a conceptual framework to investigate the dynamics of associations between serotype, serotype and serotype-specific immunity in pneumococcus populations.  We find that antigenic type (AT) is the principal determinant of non-capsular virulence factors (VF) , whereas MT is the major determinant. AT and MT are highly non-random; MT and AT are co-evolved and co-expressed. ET and CT are also found to be highly correlated, suggesting that they have synergistically adapted to a particular metabolic niche. IT and LD are found to have similar patterns of linkage disequilibrium (LD) than randomly selected genes not associated with metabolic/transport processes; AT is associated with a higher frequency of LD LD than MT LD; CT LD=0.013). CT is the first mathematical model to explain the non-overlapping association between serotypic and serotypes. TCT BC LD is a useful tool for predicting the potential impact of vaccination on the prevalence of serotypes associated with non-vaccine serotypes and for predicting how they may change under vaccination and vaccine serotype replacement.
                
                \underline{ATLAS$_{lay}$} \\
                Pneumococcal populations are highly diverse in non-antigenic genes and are commonly classified into sequence types (ST) by Multi Locus Sequence Typing (MLST) of seven metabolic housekeeping genes. STs have been documented to occur regularly throughout the past 7 decades, yet many studies (eg) show an intriguing pattern of largely non-overlapping associations between serotype and ST. It has been noted that many STs that were previously associated with vaccine serotypes now occur in association with non-vaccine serotypes. It has been proposed that a combination of immune-mediated interference between identical antigenic types and direct competition between identical metabolic types can generate non-overlapping association between antigenic and STs in populations of the bacterial pathogen Neisseria meningitidis . In this paper, we explore whether pneumococcal population structure, can be explained within a similar conceptual framework. in which pathogen strains are profiled by antigenic type, AT, metabolic type (MT) and additional non-capsular virulence factors (VF).
            \endgroup \vspace{4pt}
        }
    } 
    \caption{An case study from the eLife test set comparing summaries generated under highly lay and technical attribute values (with the length attribute being kept constant).}
    \label{fig:case_study_plos}
\end{figure*}

\label{sec:appendix}

\end{document}